# TOTAL VARIATION RECONSTRUCTION FOR COMPRESSIVE SENSING USING NONLOCAL LAGRANGIAN MULTIPLIER

*Chien Van Trinh, Khanh Quoc Dinh, Viet Anh Nguyen, and Byeungwoo Jeon*
School of Electrical and Computer Engineering, Sungkyunkwan University, Suwon, Korea
{trinhchien, dqkhanh, vietanh, and bjeon}@skku.edu

## ABSTRACT

Total variation has proved its effectiveness in solving inverse problems for compressive sensing. Besides, the nonlocal means filter used as regularization preserves texture better for recovered images, but it is quite complex to implement. In this paper, based on existence of both noise and image information in the Lagrangian multiplier, we propose a simple method in term of implementation called nonlocal Lagrangian multiplier (NLLM) in order to reduce noise and boost useful image information. Experimental results show that the proposed NLLM is superior both in subjective and objective qualities of recovered image over other recovery algorithms.

## 1. INTRODUCTION

An emerging framework of compressive sensing (CS) can reconstruct a vector $u$ of length $N$ from a measurement vector $b$ of length $M$ under certain conditions [1]. The measurement vector is sensed from an original image $u$ by:

$$b = Au \quad (1)$$

Here, $A$ is a measurement matrix. If the vector $u$ is not sparse, the matrix $A$ can be understood as being combined with a sparsifying transform. So far, many CS solutions have been developed, to name a few, Bayesian framework [2], Total variation (TV) [3-7], and Smooth Projected Landweber (SPL) [8]. Among them, TV is popularly used due to its better capability in preserving edges and boundaries compared to other recoveries [4]. It tries to optimize the constrained problem:

$$\min_{u} \|D_m u\|_p \quad \text{s.t.} \quad Au = b \quad (2)$$

where $D \in (D_x, D_y)$ denotes vertical and horizontal gradients respectively. The anisotropic or isotropic TV is differentiated by setting the value of $p$ to be 1 or 2 [4]. Because it is nonlinear and non-differentiable in terms of $\ell$p-norm [4], $D_m u = w_m$ is set and the isotropic TV in (2) is:

$$\min_{w_m, u} \|w_m\|_2 \quad \text{s.t.} \quad D_m u = w_m, Au = b \quad (3)$$

There are many solutions for TV such as Bregman method [3] and augmented Lagrangian method [4-6]. In [4], the authors demonstrated the superiority of a special augmented Lagrangian TV called Augmented Lagrangian and Alternated Direction Algorithms (TVAL3) over the other state-of-the-art algorithms. Despite of its already proven good recovery performance, similar to other TV algorithms, TVAL3 has the problem coming from regularisers seeking for a piecewise constant solution, so they tend to lose detailed information [7] even though edge objects can be preserved [4]. Furthermore, motivated by nonlocal features of natural images [9], the nonlocal means (NLM) filter is used as a regularization term called nonlocal regularization for general TV [3, 5, 7], and, particularly, for the augmented Lagrangian TV [5]. Actually, an earlier regularization comes from [3], but it is based on Bregman TV and has a slight difference compared with [5].

Generally, employing extra regularizations like the nonlocal regularization makes the optimization model more complicated. In augmented Lagrangian TV based CS recovery, the Lagrangian multiplier representing for the gradient regularization contains not only noise and image structure as well. Therefore, after updating the Lagrangian multipliers, we utilize a nonlocal means filter to reduce noise and preserve image information in the Lagrangian multiplier of gradient regularization. This method is called nonlocal Lagrangian multiplier (NLLM). Experimental results demonstrate that NLLM takes two obvious advantages: one is easy implementability and the other is reduction of over-smoothing caused by piecewise constant solution.

The rest of this paper is as following. Section 2 briefly reviews the nonlocal regularization as well as the proposed NLLM. A proposed implementation of NLLM to TV reconstruction is presented in detail in section 3. Section 4 gives experimental results, and section 5 draws some conclusions.

## 2. NONLOCAL MEANS FILTER FOR IMAGE COMPRESSIVE SENSING RECOVERY

### 2.1. Nonlocal Regularization

The method of multipliers which is investigated in [11] is widely used in image processing [4, 5, 6, 14]. For CS recovery, the augmented Lagrangian TV [4] converts the con-

This work was supported by the National Research Foundation of Korea (NRF) grant funded by the Korean government (MSIP) (No. 2011-001-7578).

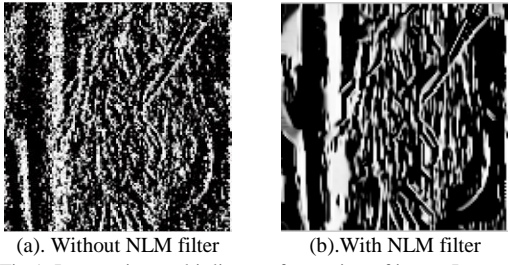

(a). Without NLM filter     (b). With NLM filter

Fig.1. Lagrangian multiplier $\upsilon_x$ of a portion of image Lena at subrate 0.3 with/without NLM

strained problem in (3) to the unconstrained minimization problem in (4) with subproblems $w_m$ and $u$; $\upsilon_m$ and $\lambda$ called Lagrangian multipliers is updated by [4, 5, 14]:

$$\upsilon_m^k = \upsilon_m^{k-1} - \beta_m \left( D_m u^k - w_m^k \right)$$
$$\lambda^k = \lambda^{k-1} - \mu \left( A u^k - b \right) \quad (5)$$

Here, $\beta_m$ and $\mu$ denote positive penalty parameters. TVAL3 shows its superiority for sparse signal as MRI images, but it is effective in preserving edges of natural image, but not for texture [5].

By the way, nonlocal regularization is proposed by Buades et al. [9] for image deblurring. There are several alternative nonlocal models applied to CS recovery of natural images such as in [3, 5]. The authors in [5] utilized the splitting technique [6] and the augmented Lagrangian TV [4] to minimize the constrained optimization problem:

$$\min_{w_m,z} \left\{ \|w_m\|_2 + \rho \|z - Fz\|_2^2 \right\} \; s.t. \; D_m u = w_m, Au = b, u = z \quad (6)$$

where $F$ is the NLM filtering operator. The constrained optimization of (6) now turns to be (7) with extra penalty parameters (i.e., $\rho$ and $\theta$) and a new Lagrangian multiplier $\gamma$ standing for the filtering operator. Besides, X. Zhang et al. also proposed another one for the Bregman TV [3]. The difficulty to take the first derivative for minimizing the $u$ subproblem is avoided by relaxation of $Fu_{k+1} \cong Fu_k$ [3]. It means that $Fu$ does not change much from previous iteration to the current one. In this paper, for fair comparison with [5], we will consider impact of nonlocal regularization and the relaxation in [3] to the augmented Lagrangian TV, so it is adapted for TVAL3 [4] as follows:

$$\min_{w_m,u} \|w_m\|_2 \; s.t. \; D_m u = w_m, Au = b, \|u - Fu\|_2 \le \varepsilon \quad (8)$$

After that, (8) is solved by the augmented Lagrangian method (i.e., the unconstrained optimization function is expressed by (9)) rather than the Bregman method [3]. Similar to [4], (9) is only optimized by separate minimization of $w_m$ and $u$ subproblems. By other words, the minimized solution for the $w_m$ subproblem can be found by the Shrinkage formula [4, 5, 14]. Thanks to the relaxation [3], two constraints $Au = b$ and $\|u - Fu\|_2$ do not need to be split into separate subproblems as in [5]. Hence, the $u$ subproblem is minimized by (9) with a gradient direction [4, 14] to reduce computation for calculating inverse matrices. Let's call this modified solver as TVNLR1 (i.e., it is distinguished from TVNLR [5]). It will be also compared with the proposed method of NLLM.

**2.2. Proposed Nonlocal Lagrangian Multiplier**

Although regularization employing the NLM filter can preserve texture of reconstructed images better, it makes the optimization function much complicated. From a viewpoint of the splitting technique [6], the method in [5] has to solve four subproblems, at each iteration, including $w_m$ (i.e., $w_x$ and $w_y$), $u$, and $z$. Moreover, the solver in [5] is more complex both in $u$ and $z$ subproblems due to existence of extra variable of $z$ related to the nonlocal regularization. In fact, minimization of the z subproblem increases computational complexity due to high cost of the nonlocal means filter. Finally, the extra Lagrangian multiplier $\gamma$ representing filtering the image $u$ needs to be controlled.

In comparison with the method [5], the solver based on [3] for the augmented Lagrangian TV looks simpler than the solution proposed in [5] because the nonlocal regularization does not need to split into a separate subproblem. Otherwise, compared to the original solver [4], this work still needs to solve the $u$ subproblem which increases complexity. Therefore, a new approach based some other prior information is investigated in this paper to utilize the NLM filter with lower cost.

In the augmented Lagrangian method, penalty parameters and Lagrangian multipliers are key factors to solve convex optimization problems. Motivated by the importance of them, the authors in [10] proposed a scheme to update the penalty parameters according to the value of the Lagrangian multipliers so that optimization quality is improved compared to the conventional method [11]. However, for CS recovery, based on prior image information in the Lagrangian multiplier $\upsilon_m$, we exploit the NLM filter to update $\upsilon_m$.

It is clear that, from (4), the Lagrangian multipliers $\upsilon_m$ and $\lambda$ are matrices respectively representing the gradient images and the measurement vector $b$. Specially, the Lagrangian multiplier $\upsilon_m$ will be updated from the gradient image $D_m u$. Hence, $\upsilon_m$ is also an erroneous version of gradient image

$$\min_{w_m,u} \left\{ \left( \sum_{m \in (x,y)} \left( \|w_m\|_2 - \upsilon_m^T (D_m u - w_m) + (\beta_m/2)\|D_m u - w_m\|_2^2 \right) \right) - \lambda^T (Au - b) + \frac{\mu}{2}\|Au - b\|_2^2 \right\} \quad (4)$$

$$\min_{w_m,u,z} \left\{ \left( \sum_{m \in (x,y)} \|w_m\|_2 - \upsilon_m^T (D_m u - w_m) + (\beta_m/2)\|D_m u - w_m\|_2^2 \right) - \lambda^T (Au - b) + \frac{\mu}{2}\|Au - b\|_2^2 + \rho \|z - Fz\|_2^2 - \gamma^T (u - z) + \frac{\theta}{2}\|u - z\|_2^2 \right\} \quad (7)$$

$$\min_{w_m,u} \left\{ \left( \sum_{m \in (x,y)} \|w_m\|_2 - \upsilon_m^T (D_m u - w_m) + (\beta_m/2)\|D_m u - w_m\|_2^2 \right) - \lambda^T (Au - b) + \frac{\mu}{2}\|Au - b\|_2^2 - \gamma^T (u - Fu) + \frac{\theta}{2}\|u - Fu\|_2^2 \right\} \quad (9)$$

Table 1: Augmented Lagrangian TV reconstruction using NLLM

| |
|---|
| **Input**: Measurement matrix $A$, measurement vector $b$, Lagrangian multipliers and penalty parameters, $u_0 = A^T b$ |
| **While** Outer stopping criteria unsatisfied **do** |
|     **While** Inner stopping criteria unsatisfied **do** |
|         Solve $w_m$ subproblem by computing (11) |
|         Solve u subproblem by computing (13) with estimation of gradient direction via (14) |
|     **End** |
|     Update multiplier $\upsilon_m$ using NLM filter by (15) |
|     Update multiplier $\lambda$ by (16) |
| **End** |
| **Output**: The final CS recovered image. |

$D_m u$ containing lots of noise if it is updated by the traditional method (by (5)) as illustrated in Fig. 1(a). Furthermore, according to the splitting technique [6], $\upsilon_m$ takes part in solving the $w_m$ subproblem as in (4). $\upsilon_m$ also plays a role in solving the $u$ subproblem (i.e. see (12)). More exact $\upsilon_m$ will provide more accurate solution to the $w_m$ subproblem and better reconstructed image $u$. Consequently, a proper change of $\upsilon_m$ is expected to enrich the quality of the augmented Lagrangian TV algorithm. Much noise contained in $\upsilon_m$ gives rise to less structural information as visualized in Fig. 1(a). Hence, it naturally calls for necessity of noise reduction of $\upsilon_m$ for enhanced structures. In this regards, some lowpass filters as Wiener filter or Gaussian filter are conceivable, but they can easily make the reconstructed image over-smoothed [14]. Instead, the NLM filter is applied, by noting its strengths in denoising and preserving textures in images, to $\upsilon_m$ to form a new update for the Lagrangian multiplier as:

$$\hat{\upsilon}_m^k = F \upsilon_m^k$$
$$\upsilon_m^k = \upsilon_m^{k-1} - \beta_m \left( D_m u^k - w_m^k \right)$$

The NLM filter is used after updating the Lagrangian multiplier by (5), and Fig. 1(b) backs up its effectiveness. Thanks to the NLM filter, noise hardly occurs in $\upsilon_x$, thus edge objects and details of Lena's hair are much clearer if compared with a result by traditional method [4] as shown in Fig. 1(a).

## 3. AUGMENTED LAGRANGIAN TV USING NLLM

### 3.1. Implement NLLM to Augmented Lagrangian TV

The effectiveness of the proposed method is evaluated using augmented Lagrangian TV [4, 14]. Since it is difficult to directly minimize the cost function of (4) with both $w_m$ and $u$ at the same time, the splitting technique [4, 6] is used to separate the cost function into subproblems of $w_m$ and $u$. It means that, at each iteration, (4) minimizes the subproblems of $w_m$ and $u$ separately under an assumption that a solution of the other subproblem is available.

**$w_m$ subproblem:** Given $u$, the optimization problem associated with $w_m$ can be expressed by:

$$\min_{w_m} \left( \sum_{m \in (x,y)} \|w_m\|_2 - \upsilon_m^T (D_m u - w_m) + (\beta_m / 2) \|D_m u - w_m\|_2^2 \right) \quad (10)$$

The closed form of (10) is formulated by the Shrinkage formula [4, 14] with $\odot$ denoting an element-wise product:

$$w_m = \max \left\{ \left\| D_m u - \frac{\upsilon_m}{\beta_m} \right\|_2 - \frac{1}{\beta_m}, 0 \right\} \odot \frac{(D_m u - \upsilon_m / \beta_m)}{\|D_m u - \upsilon_m / \beta_m\|_2} \quad (11)$$

**u subproblem:** Given $w_m$, the u subproblem is equivalent to:

$$\min_u \left\{ \begin{array}{l} \sum_{m \in (x,y)} -\upsilon_m^T (D_m u - w_m) + (\beta_m / 2) \|D_m u - w_m\|_2^2 \\ -\lambda^T (Au - b) + (\mu / 2) \|Au - b\|_2^2 \end{array} \right\} \quad (12)$$

The solution $\hat{u}$ in the minimization of (13) is sought by using the steepest descent with the Barzilai−Borwein step $\alpha$ [4]:

$$\hat{u} = u - \alpha d \quad (13)$$

where $d$ stands for the gradient direction of the object function of (14), and is calculated by:

$$d = \sum_{m \in (x,y)} D_m^T (\beta D_m u - w_m - \upsilon_m) + \mu A^T (Au - b) - A^T \lambda \quad (14)$$

The Lagrangian multipliers are updated as follows:

$$\hat{\upsilon}_m^k = F \upsilon_m^k$$
$$\upsilon_m^k = \upsilon_m^{k-1} - \beta_m \left( D_m u^k - w_m^k \right) \quad (15)$$
$$\lambda^k = \lambda^{k-1} - \mu \left( Au^k - b \right) \quad (16)$$

The proposed update of $\upsilon_m$ is integrated into the augmented Lagrangian TV as briefly depicted by Table 1.

### 3.2. Discussion

Recently, researchers [12, 13] have investigated about reducing error in gradient domain for denoising. The authors [12] confirm that application of the bilateral filter for gradient image attains good quality of denoised images. The gradient matching method proposed in [13] shows excellent ability in preserving texture of denoised images. Additionally, for CS recovery, theorem 2 in [15] states that the spatial error is bounded by the gradient error. It means that the smaller gradient error results in the smaller error in spatial domain. With all analyses above, using NLLM (i.e., the Lagrangian multiplier $\upsilon_m$ is considered as an erroneous version of gradient image) promises better performance than that of using the NLM filter as nonlocal regularizations [3, 5] working with error in spatial domain.

Compared with TVAL3 in [4], CS recoveries employing TVNLR [5], TVNLR1 [3], and NLLM increase computational complexity due to the computational cost of NLM filter [14]. More clearly, if the search range and size of similarity patches of NLM filter are $[-S, S]^2$ and $(2W + 1)(2W + 1)$, respectively, then, with an image of size $N \times N$, the computational complexity of this filter is $O(N^2(2S + 1)^2(2W + 1)^2)$. Obviously, the NLM filter takes high cost due to the large size of natural images. Both TVNLR [5] and TVNLR1 [3] employ the NLM filter to solve the subproblems, so they suffer much from burden of the NLM filter. By contrast, NLLM only calls the NLM filter if the $u$ and $w_m$ subproblems are solved to satisfy the inner stopping criteria [4], so it is expected to lower cost than that of using the NLM filter as regularization. One more advantage of NLLM compared with TVNLR [5] and TVNLR1 [3] is that it does not increase the optimization complexity of subproblems compared with the original TVAL3 [4].

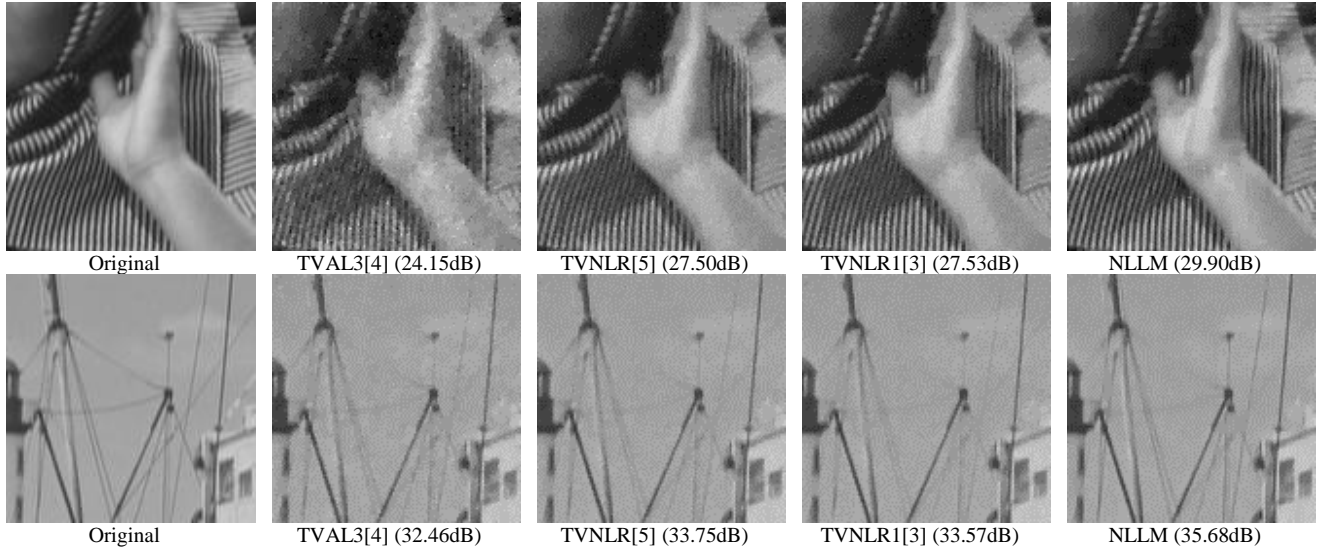

Fig. 2 Visual quality comparison of cropped Barbara and Boats recovered by different recoveries

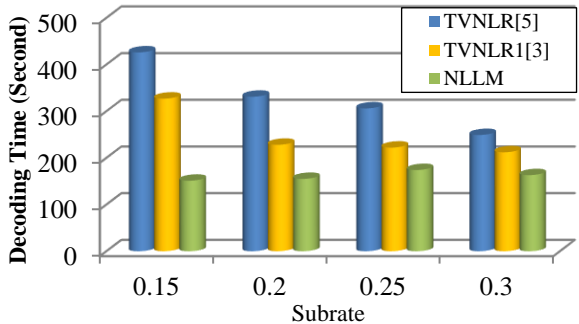

Fig. 3 Comparison of decoding times (image Boats)

## 4. EXPERIMENTAL RESULTS

Performance of recovery algorithms is evaluated by using natural images of size 256x256. In our experiments, CS measurements are acquired using a Gaussian random matrix in [5]. The penalty parameters are empirically selected to obtain the best quality of recovered images in terms of PSNR ($\mu=512$, $\beta_m=32$, $\rho=1$, $\theta=4$, $\alpha=1$). The searching window size of the NLM filter is 13x13, whilst its neighborhood is of size 7x7. The filtering parameter is set to 0.03 if the NLM filter is used as nonlocal regularization [5], and it is set to 0.19 for NLLM. The simulation uses MatlabR2011a running on a desktop Intel Core i3, RAM 4G.

Table 2 indicates the quality of images recovered by methods using the NLM filter composing of TVNLR [5], TVNLR1 [3], and NLLM compared with TVAL3 [4]. It is obvious to see that all proposed methods TVNLR [5], TVNLR1 [3], and NLLM improve the objective quality of recovered images against the original TVAL3 [4] for all tested images. The PSNR improvement over TVAL3 [4] is up to 2.71dB, 2.73dB and 4.49dB for TVNLR [5], TVNLR1 [3], and NLLM, respectively. Among the three proposed methods, NLLM which is the simplest in terms of implementation achieves notable increments in recovered image quality. On average of four tested images, NLLM gains PSNR of by about 2.66dB compared with TVAL3 [4]. Although TVNLR1 [3] and TVNLR [5] are more complex than NLLM, the average PSNR gain is only by about 1.31dB and 1.51dB, respectively. Specially, for the image Boats which contains weak edge objects, NLLM method is better than TVNLR [5] by up to 1.22dB. The method applying the NLM filter as regularization seems to be only slightly efficient for image Monarch containing blurred objects, whilst NLLM significantly improves recovered image quality.

Comparisons in PSNR of the proposed method with some CS recoveries of tree-structured CS with variational Bayesian analysis using Discrete Cosine Transform (TSDCT) [2], SPL using Discrete Wavelet Transform (SPLDWT) [8], and SPL using Dual-tree Discrete Wavelet Transform (SPLDDWT) [8] are also shown in Table 3. Thanks to the NLM filter, noise is well suppressed in the Lagrangian multiplier $\upsilon_m$, so NLLM has better quality of reconstructed images than the other methods. Compared with TSDCT [2], our method outperform by up to 7.36dB (image Monarch at subrate 0.3). Furthermore, NLLM is not only better than TSDCT [2], but also than SPLDWT [8]. In the best case (i.e., with image Leaves at subrate 0.3), PSNR gain by NLLM is about 6.88dB. As shown in Table 3, the SPLDDWT [8] provides the second best performance in most of the considered subrates and images, but NLLM also has PSNR gain by up to 7.04dB with image Leaves.

Table 2: Performance comparison of nonlocal regularizations and the proposed method (PSNR: dB)

| Subrate | | 0.15 | 0.2 | 0.25 | 0.3 |
|---|---|---|---|---|---|
| Barbara | TVAL3[4] | 23.31 | 24.23 | 25.03 | 26.08 |
| | TVNLR [5] | **24.46** | 25.93 | 27.42 | 28.63 |
| | TVNLR1 [3] | 24.26 | 25.79 | 27.27 | 28.46 |
| | NLLM | 24.30 | **26.42** | **28.72** | **30.57** |
| Leaves | TVAL3[4] | 21.17 | 23.18 | 24.99 | 26.77 |
| | TVNLR [5] | 23.34 | 25.68 | 27.70 | 29.26 |
| | TVNLR1 [3] | 23.17 | 25.52 | 27.72 | 29.20 |
| | NLLM | **24.35** | **26.81** | **28.72** | **30.27** |
| Monarch | TVAL3[4] | 26.48 | 28.21 | 30.10 | 31.57 |
| | TVNLR [5] | 26.93 | 28.65 | 30.06 | 31.52 |
| | TVNLR1 [3] | 26.77 | 28.46 | 29.61 | 31.22 |
| | NLLM | **28.40** | **30.08** | **31.71** | **32.98** |
| Boats | TVAL3[4] | 26.96 | 28.55 | 29.89 | 31.14 |
| | TVNLR [5] | 28.36 | 29.92 | 31.34 | 32.54 |
| | TVNLR1 [3] | 28.16 | 29.73 | 31.04 | 32.31 |
| | NLLM | **29.45** | **31.06** | **32.69** | **33.76** |

Fig. 2 compares subjective quality of reconstructed images of cropped Barbara and Boat at subrate 0.3. The recovered image Barbara demonstrates that NLLM achieves significant improvement in subjective quality. It better preserves details of objects such as scarf lines. Moreover, NLLM can recover some weak edge objects on the image Boats better than others.

The decoding time of three TV recoveries using the NLM filter (i.e., TVNLR [5], TVNLR1 [3], NLLM) is shown in Fig. 3 for image Boats. At all considered subrates, TVNLR [5] spends the longest computational time (e.g., at subrate 0.15, the decoding time is 426s) to recover images because it needs to solve 4 subproblems, while at the same subrate, TVNLR1 spends 327s to solve the 3 subproblems. Moreover, the NLM filter should be applied in per iteration since TVNLR [5] and TVNLR1 [3] use it as regularization. As a result, these algorithms suffer much from the computational complexity of NLM filter. In contrast, NLLM only uses the NLM filter if having sufficient change of $u$ and $w_m$ subproblems, it spends the least decoding of 151s. Roughly speaking, TVNLR [5] and TVNLR1 [3] averagely take more decoding time than NLLM by 3 and 2 times, respectively.

## 5. CONCLUSION

This paper proposes a new approach to update the Lagrangian multiplier for TV reconstruction called Nonlocal Lagrangian multiplier (NLLM). Different from TV recovery using the NLM filter as regularization for optimization function, NLLM utilizes the NLM filter to update the Lagrangian multiplier. Although NLLM is much simpler to implement than that using the nonlocal means filter as regularization in terms of implementation, its experimental results demonstrate significant improvement of recovered images both in subjective and objective qualities.

Table 3: Performance comparison of various CS recoveries (PSNR: dB)

| | Subrate | 0.15 | 0.2 | 0.25 | 0.3 |
|---|---|---|---|---|---|
| Barbara | TSDCT [2] | 22.79 | 23.57 | 24.61 | 25.57 |
| | SPLDWT [8] | 23.35 | 23.96 | 24.74 | 25.42 |
| | SPLDDWT[8] | 23.63 | 24.32 | 25.00 | 25.67 |
| | NLLM | **24.30** | **26.42** | **28.72** | **30.57** |
| Leaves | TSDCT [2] | 19.02 | 20.81 | 22.13 | 23.39 |
| | SPLDWT [8] | 19.56 | 20.99 | 21.99 | 22.86 |
| | SPLDDWT[8] | 19.98 | 21.37 | 22.36 | 23.22 |
| | NLLM | **24.35** | **26.81** | **28.72** | **30.27** |
| Monarch | TSDCT [2] | 21.62 | 23.39 | 24.55 | 25.62 |
| | SPLDWT [8] | 23.11 | 24.72 | 25.90 | 27.19 |
| | SPLDDWT[8] | 23.68 | 25.26 | 26.41 | 27.80 |
| | NLLM | **28.40** | **30.08** | **31.71** | **32.98** |
| Boats | TSDCT [2] | 24.84 | 26.33 | 27.61 | 28.97 |
| | SPLDWT [8] | 25.73 | 26.79 | 27.84 | 28.81 |
| | SPLDDWT[8] | 25.99 | 27.02 | 28.05 | 29.02 |
| | NLLM | **29.45** | **31.06** | **32.69** | **33.76** |